\documentclass{article}

\usepackage{arxiv}
\usepackage[utf8]{inputenc} 
\usepackage[T1]{fontenc}    
\usepackage{hyperref}       
\usepackage{url}            
\usepackage{booktabs}       
\usepackage{amsfonts}       
\usepackage{nicefrac}       
\usepackage{microtype}      
\usepackage{lipsum}
\usepackage{graphicx}
\usepackage{multicol}
\usepackage{upgreek}

\title{Intweetive Text Summarization}

\author{
  Jean Valère Cossu\\
   Laboratoire Informatique d'Avignon - Université d'Avignon \\
   Avignon Cedex 9, France \\
  \texttt{jvcossu@gmail.com} \\
   \AND
   Juan-Manuel Torres-Moreno \\
   Laboratoire Informatique d'Avignon - Université d'Avignon \\
   Avignon Cedex 9, France and \\
   Polytechnique Montréal, CP 6079, succ. Centre-ville, Montréal (Québec) Canada H3C3A7\\
   \texttt{juan-manuel.torres@univ-avignon.fr} \\
   \And
  Eric SanJuan\\
   Laboratoire Informatique d'Avignon - Université d'Avignon \\
   Avignon Cedex 9, France \\
  \texttt{eric.sanjuan@univ-avignon.fr} \\
  \And
  Marc EL-Bèze\\
   Laboratoire Informatique d'Avignon - Université d'Avignon \\
   Avignon Cedex 9, France \\
  \texttt{marc.elbeze@univ-avignon.fr} \\
}

\begin{document}

\maketitle

\begin{abstract}
The amount of user generated contents from various social medias allows analyst to handle a wide view of conversations on several topics related to their business.
Nevertheless keeping up-to-date with this amount of information is not humanly feasible. 
Automatic Summarization then provides an interesting mean to digest the dynamics and the mass volume of contents.
In this paper, we address the issue of tweets summarization which remains scarcely explored.
We propose to automatically generated summaries of Micro-Blogs conversations dealing with public figures E-Reputation.
These summaries are generated using key-word queries or sample tweet and offer a focused view of the whole Micro-Blog 	network.
Since state-of-the-art is lacking on this point we conduct and evaluate our experiments over the multilingual CLEF RepLab Topic-Detection dataset according to an experimental evaluation process. 
\end{abstract}

\section{Introduction}
\label{sec:intro}

The amount of the information collectively generated by users on online social networks have drastically increased during these last years. 
And Online social interactions provide a real-time reflect of real-world events on people opinions but frequently remains unprocessed due to the sheer quantities of it.
Hence understanding social events become crucial for entities concerned with their online reputation.
These entities typically spend a lot of money to obtain reliable satisfaction polls using call centers and surveys, and online social networks are certainly carrying key information to anticipate and react to the versatility of public opinions.
The high volume of conversations and the velocity of answers make impossible to keep up-to-date with all events of interests related to a specific business. 
Considering this amount of documents, automatic approaches are needed. 
More precisely Automatic text summarization appears indispensable to cope with the increasing amount of information \cite{torres:2011,torres:2014}.  

An increasing number of Social Media Analysis (SMA) services and community managers use Micro-Blog streams to analyze keep in touch with the market mood.
Twitter offers the Trends service and if Social Event detection itself has already been studied it does not end story. 
Indeed, it does only provide cues of the existence of these events and an access to the 1,500 last related tweets either they are really informative or not.
As well as usual generic summarization will always provide the same global abstract from a pool of tweets.
End-Users are then not aware to capture the main information to completely understand what this event is about.
And probably the most crucial aspect, what are Twitter-users saying about this event.

Towards this goal, we are interested in using these events as queries to produce guided automatic summaries of the tweets streams.
Considered queries could be a topic, conceptual topic, a cluster label or even a sample tweet illustrating the topic.
The main objective is to allow business analyst to quickly digest the focused information avoiding the peculiar choice of keywords and concepts definition which is non accessible to non-technical person.

This problem can also be considered as a Question-Answering (QA) issue where system has to exactly answer a question expressed in Natural Language. 
QA systems are confronted with a fine and difficult task because they are expected to supply specific information and not whole documents or pools of documents. 
Currently there exists a strong demand for this kind of text processing systems on the Internet. 
A QA system comprises, \emph{a priori}, the following stages \cite{JAC00}:
\begin{itemize}
 \item Transform the questions into queries, then associate them to a set of documents;
 \item Filter and sort these documents to compute various degrees of similarity;
 \item Identify the sentences which might contain the answers, then extract text fragments from those that constitute the answers. 
 In this phase an analysis using Named Entities (NE) is essential to find the expected answers.
\end{itemize}

Search engine are more about answering queries whatever their form while in our case topic label can be very specific and sometimes words from the queries are even absent from the contents when the query is about '\textit{Ethics}' or '\textit{Innovation}'.
As such, we conduct an experimental evaluation for query-driven Micro-Blogs summarization. 
We investigate 2 different statistical summarization systems evaluated with FRESA for summarization informativity and MAP for the ranking and information retrieval aspect.

The rest of this paper is organized as follows: Section~\ref{sec:state} provides related works in Automatic Summarization and Micro-Blogs Summarization. In Section~\ref{sec:app}, we discuss the issue and provide details about our approaches. A discussion of our results is provided in Section~\ref{sec:results}. Finally, Section~\ref{sec:conclu} gives our conclusions on our work and open several perspectives.

\section{Related work}
\label{sec:state}

An abstract is, by far, the most concrete and most recognized kind of text condensation \cite{ANS79,torres:2011}.
We adopted a simple method, usually called \textit{extraction}, that allows to generate summaries by extracting 'relevant' sentences \cite{torres:2011,luhn1958,edmundson1969,IND99}. 
Essentially, this extraction aims at producing a shorter version of the text by selecting the most relevant sentences of the original text, which we juxtapose without any modification.
The vector space model \cite{SAL71,SAL83} has been used in information extraction, Information Retrieval (IR), QA, and it may also be used in text summarization \cite{dacunha:07}.

Most research efforts in summarization emphasize generic summarization \cite{Abracos&Lopes97,Teufel&Moens97,Hovy&Lin99,torres:2014}.
User query terms are commonly used in IR tasks.
However, there are few papers in literature that propose to employ this approach in summarization systems \cite{Kupiec&al.95,Tombros&al.98,SCH01}.
In the systems described in \cite{Kupiec&al.95}, a learning approach is performed.
A document set is used to train a classifier that estimates the probability that a given sentence has to be included in the extract. 

In \cite{Tombros&al.98}, several features (document title, location of a sentence in the document, cluster of significant words and occurrence of terms present in the query) are applied to score the sentences.
In \cite{SCH01} learning and feature approaches are combined in a two-step system: a training system and a generator system. 
Score features include short length sentence, sentence position in the document, sentence position in the paragraph, and \textit{TF.IDF} metrics. 
Recent works integrate more sophisticated methods from Machine Learning (ML) and Natural Language Processing (NLP) fields. 

Automatically summarizing Micro-Blogs' conversation is a relatively new area of research. 
It recently attracted several research teams in Europe used to focus on automatic summarization of Events. 
With their dataset on tweet-sized topic-specific summaries~\cite{sharifi2010experiments} became a reference to benchmark Micro-Blogs summarization approaches.
Which by the way also inspired~\cite{zubiaga2012summarization} who focused on soccer-match summarization.
Partly using the same datasets, Mackie \textit{et al.}~\cite{mackie2014comparing,mackie2014choosing} proposed a comparative evaluation of different summarization methods and looked for the most effective evaluation metrics from these summarization methods.

These works intended to be representative but their evaluation was event-detection centered since their purpose was to find the most relevant tweet with regards to the given event.
Although detecting goals with the mass of tweets on a specific topic in a short period of the timeline appears to be solved~\cite{zubiaga2012summarization} producing a summary of the '\textit{Defense efficiency}' during the match, which is nearer usual Micro-Blogs summarization scenarios, is still a wile open door.

Summarizing Micro-Blogs can be viewed as an instance of automated text summarization which is the problem of automatically generating a condensed version of the most important content from a pool of contents (a long document or several documents). 
This summary can be generated for a particular user or to answer a specific question, this is called guided or personalized summarization. 
Having as purpose to group tweets regarding their usefulness to a given topic, Wen and Marshall~\cite{wen2014automatic} proposed to digest Twitter trending topics using Hidden Markov Models to rank the most important tweets related to each of their selected topics.

\section{Problem statement and Selected Approaches}
\label{sec:app}

We define the Micro-Blogs summarization issue as the task that, given a pool of tweets and a user generated query, provides a focused view of the pool.
The query is a real-world complex question (called long query) answering, in which the answer is a summary constructed from a set of relevant documents.

In~\cite{zubiaga2012summarization} the authors also focused on event detection. In our case the selected events are given by the user and its impact is two-fold:
\begin{itemize}
 \item First, the user could give a non relevant query which means there are no tweets (or only one tweet) from the pool that answer his query, or a too generic query which in this case can be answered by a too larger set of tweets;
 \item Secondly, a given tweet could answer two different queries (and this information may not be in the reference); 
\end{itemize}

Our objective is to use summarizers as tweets-selector. 
The systems are provided tweets from those they have to chose the most representative ones that answer the query but also contains information from the complete pool. 
So the user is able to understand the whys and wherefores.

Below we compare two tweets selection methods and we test them with a set of provided queries.

Our generic summarization system includes a set of eleven independent metrics combined by a Decision Algorithm. 
Query-based summaries can be generated by our systems using a modification of the scoring method. 
In both cases, no training phase is necessary in our system.

\subsection{Cortex Summarizer}

Cortex \cite{cortex,DBLP:journals/corr/abs-0905-2990} is a single-document extract summarization system.
It uses an optimal decision algorithm that combines several metrics. 
These metrics result from processing statistical and informational algorithms on the document vector space representation. 

The idea is to represent the text in an appropriate vectorial space and apply numeric treatments to it.
In order to reduce complexity, a pre-processing is performed on the question and the document: words are filtered, lemmatized and stemmed. 

The Cortex system uses 11 metrics (see \cite{metrics,DBLP:journals/corr/abs-0905-2990} for a detailed description of these metrics) to evaluate the sentence's relevance. 
For instance, the topic-sentence overlap measure assigns a higher ranking for the sentences containing question words and makes selected sentences more relevant. 
The overlap is defined as the normalized cardinality of the intersection between the query word set $T$ and the sentence word set $S$. 
\begin{eqnarray}
\textrm{Overlap}(T,S)=\frac{card(S \cap T)}{card(T)}
\end{eqnarray}

The system scores each sentence with a decision algorithm that relies on the normalized metrics. 
Before combining the votes of the metrics, these have been split into two sets: one set contains every metric $\lambda^i > 0.5$, while the other set contains every metric $\lambda^{i} < 0.5$ (values equal to 0.5 are ignored). 
We then compute two values $\alpha$ and $\beta$, which give the sum of distances (positive for $\alpha$ and negative for $\beta$) to the threshold 0.5 (the number of metrics is $\Gamma$, which is 11 in our experiment):
\begin{eqnarray}
\alpha = \sum_{i=1}^\Gamma (\lambda^i - 0.5) ;\ \ \lambda^i > 0.5   \\
\beta = \sum_{i=1}^\Gamma (0.5 - \lambda^i) ;\ \ \lambda^i < 0.5 
\end{eqnarray}

The value given to each sentence $s$ given a query $q$ is calculated with:
\begin{center}
\begin{equation}
\begin{array}{l}
\mbox{if} (\alpha > \beta) \\
~~~~~~~~\mbox{then}~~ \mbox{Score}(s,q) = 0.5 + \frac{\alpha}{\Gamma} \\
~~~~~~~~\mbox{else}~~ \mbox{Score}(s,q) = 0.5 - \frac{\beta}{\Gamma} \\
\end{array}
\end{equation}
\end{center}

The Cortex system is applied to each document of a topic and the summary is generated by concatenating higher score sentences.

\subsection{Artex}
\label{sec:artex}


\textsc{Artex}~\cite{torres2012artex} computes the score of each sentence by calculating the inner product between a sentence vector, an \textsl{average pseudo-sentence} vector (the ``global topic'') and an \textsl{average pseudo-word} vector (the ``lexical weight'').
The summary is generated concatenating the sentences with the highest scores.

An average document vector which represents the ``global topic'' of all sentences vectors is constructed.
The ``lexical weight'' for each sentence, i.e. the number of words in the sentence, is obtained.
A score for each sentence is calculated using their proximity with the ``global topic'' and their ``lexical weight''. 
Let $\vec{s}_{\mu}=(s_{\mu,1},s_{\mu,2},\ldots,s_{\mu,N})$ be a vector of the sentence $\mu=1,2,\ldots,\uprho$. 
The \textsl{average pseudo-word} vector ${\vec{a}}=[a_\mu]$, was defined as the average number of occurrences of $N$ words used in the sentence $\vec{s}_\mu$:
\begin{equation}
	a_\mu =  \frac{1}{N} \sum_j s_{\mu,j} 
\end{equation}
\noindent and the \textsl{average pseudo-sentence} vector ${\vec{b}}=[b_j]$ as the average number of occurrences of each word $j$ used through the $\uprho$ sentences: 
\begin{equation}
	b_j =  \frac{1}{\uprho} \sum_\mu s_{\mu,j} 
\end{equation}
The weight of each sentence is calculated as follows:
\begin{equation}
	\omega(\vec{s}) = \left( \vec{s} \times \vec{b} \right) \ \times \vec{a} 
\label{eq:artex}
\end{equation}

Finally, a random summarizer was implemented in order to create a baseline system.
This baseline system picks 15 sentences at random from the tweet set.
A random value in the range $[0,1]$ was assigned to each sentence $i$.
\section{Experimental Setup and Evaluation}
\label{sec:results}

In this section, we describe the experimental evaluation that we undertake to demonstrate the effectiveness of the proposed method to address the following question:
\begin{enumerate}
\item For a given query, are automatic summarizers able to  retrieve (and return as summary) tweets which were manually selected as answering this query ?
\item How much the returned tweets are informative, regarding the expect set of tweets and the complete pool ?
\end{enumerate}

We use the RepLab'2013 Topic Detection task where we are given a set of tweets concerning each entity and experts queries to evaluate our proposal.
That is to say: given a query to propose an overview of entity's e-Reputation using provided Micro-Blog contents. 

\subsection{Replab Collection}

We conducted our evaluation using the RepLab bilingual collection which includes more than 140,000 tweets spread in 9,750 clusters. 
The collection covers a set of 61 entities from 4 economic domains: Automotive, Artists, Banking and Universities.
More details about the collection and its annotation are available in~\cite{amigo2013overview,carrillo2014orma}\footnote{Data are available at \url{http://nlp.uned.es/replab2013/}}.

We will consider the clusters as queries for the rest of this paper.
For each entity around 2,200 tweets are provided with a set of 160 queries (with around 18 tweets per query).
Queries address a large set of facts: events, conversations, songs or products comments, news reports etc.. 
We dismissed all queries having more than 20 related tweets (such as trash called 'Other topics') and less than 2 tweets (in this case the query should be the tweet) which leads to a total number of queries of 3,521.
We kept queries with very similar labels such as 'Pictures on Social Network' and 'Photo in Social Network'.
Note that all queries are in English even if all answering tweets are in Spanish.

\subsection{Evaluation Protocol}

Evaluating summaries is known to be a challenging part~\cite{mackie2014comparing,mackie2014choosing}. 
In this work, the reference summaries for the systems summaries to evaluate against are the clusters defined by the specialists from Llorente \& Cuenca\footnote{\url{http://www.llorenteycuenca.com/}}, a leading Spanish E-Reputation firm.
Clusters are groups of tweets that are related by a common theme.
They represent a focused view of the entity E-Reputation regarding this theme.

We considered a 2 step evaluation. 
First we evaluate the quality of the tweets hierarchy where we expect our system the return as $N$ first tweets those that were tagged by the experts as belonging to the cluster labeled with the query. 
Ranking quality is estimated separately for each query using the \textit{Mean Average Precision} (MAP). 
MAP allows comparing an ordered vector (output of a submitted method) to a binary reference (manually annotated data). 
The MAP is computed as follows:

\begin{equation}
\label{map}
	MAP = \frac 1n  \sum_{i=1}^{N} p(i) R(i)
\end{equation}

\noindent where $N$ is the total number of tweets, $n$ the number of tweets correctly found (i.e. true positives), $p(i)$ the precision at rank $i$ (i.e. when considering the first $i$ tweets found) and $R(i)$ is $1$ if the $i^{th}$ tweets is related to the query, and 0 otherwise.
Then based on the hierarchy we select the 15 first sentences to compose the summary.
In order to evaluate the quality of the generated summaries we compare our proposed summaries to the reference summary using FRESA. 

FRESA (\textit{A FRamework for Evaluating Summaries Automatically}) is a method based on information theory which evaluates summaries without using human references \cite{torres:10poli}. 
\textsc{Fresa} is based on the works of \cite{lin:06,louis-nenkova:2009:EMNLP}. 
\textsc{Fresa} calculates the divergence between a source text $P$ and a summary $Q$ ,$\mathcal{D}(P||Q)$ via n-grams statistics (1-grams, 2-grams, SU4-grams and their combinations)\footnote{FRESA may be donwloaded from \url{http://fresa.talne.eu/}}. 
We also propose to compute FRESA between the automatic summaries (and reference summary), and the initial set of document to estimate an informativity.

\subsection{Performances}
The main goal of our evaluation was to verify if Automatic Summarizers are efficient enough to provide a reduced set of tweets which is informative enough regarding a given query.
Table \ref{tab:fresa1} shows the FRESA score between the summarizers and the source documents which is the total pool of tweets. 
The way the RepLab was built lead to very scores when it comes to compare the manual references for a given query to the complete pool of tweets. 
Clusters where made of tweets that are only focusing on a specific event without concerns to the rest of the pool while summarizer tried to include a part of background information. 

\begin{table}[!ht]
\centering
\begin{tabular}{|c|c|c|c|c}
	\hline
	\multicolumn{4}{c}{\bf Summarizer}\\
	\hline
	\bf Reference &\sc {\sc Cortex} & \textsc{\textsc{Artex}} & \sc Baseline  \\
	\hline
	0.0022 &\bf 0.009   & 0.007    	& 0.006   \\
	\hline
\end{tabular}
\caption{FRESA results: Summarizers vs Test corpora.}
\label{tab:fresa1}
\end{table}

During the annotation process, tweets were grouped in 'clusters' (with a various size from 1 to 150) and then annotators selected a label from the cluster which represent the main concept (the query) expressed in the cluster. 
On complicate queries where many misunderstandings can take place, once the process was done, if some tweets were more relevant regarding another cluster, they were switched to this new cluster.
Table \ref{tab:fresa2} shows the FRESA score between the summarizers and the Human reference corpora.

\begin{table}[!ht]
\centering
\begin{tabular}{|c|c|c|c}
	\hline
	\multicolumn{3}{c}{\bf Summarizer}\\
	\hline
	\sc {\sc Cortex} & \textsc{\textsc{Artex}} & \sc Baseline  \\
	\hline
	 0.059  &\bf 0.146  	& 0.048  \\
	\hline
\end{tabular}
\caption{FRESA results: Summarizers vs Human Reference corpora.}
\label{tab:fresa2}
\end{table}

In order to see which of the tweets selected by summarizers are relevant, we computed the MAP to compare summarizers ranking to the Human Reference. 
Table \ref{tab:map} shows the Average MAP obtained by each summarizer. 
Randomly choosing tweets and affecting them weights obviously perform really bad with regards to the MAP since there is almost no chance to put the expected tweets at the heading ranks.

\begin{table}[!ht]
\centering
\begin{tabular}{|c|c|c|c}
	\hline
	\multicolumn{3}{c}{\bf Summarizer}\\
	\hline
	\sc {\sc Cortex} & \textsc{\textsc{Artex}} & \sc Baseline  \\
	\hline
	 0.0040   &\bf 0.116   	& 0.001   \\
	\hline
\end{tabular}
\caption{MAP results: Summarizers output ranking vs Human Reference.}
\label{tab:map}
\end{table}

We consider a query which all systems reported at least one relevant tweets: 'Annie Le's Family Sues Yale' this query refers to a student that was murdered on the campus of the Yale University. MAP obtained by the systems are 0.3571, 0.0714, 0.0179 (respectively for 'Artex', 'Cortex' and the baseline). FRESA values for this particular query are 0.48111, 0.19042 and 0.10028.
The summarizers (expecting the baseline) returned the following content:
\textit{'Report: Annie Le’s Family Sues Yale University After Grad Student’s Killing: NEW HAVEN, CT - The family of a slain Ya...' } as most relevant tweet, mainly because the query itself was extracted from the tweets. Then if 'Artex' also extracted more tweets related to this event, 'Cortex' found tweets concerning other campus matters.

Table \ref{tab:map2} shows the number of query for which summarizer had an a non-null MAP that is to say, for this queries systems were able to select at least one relevant within the 5-top documents returned. 
The baseline low MAP performance is also illustrated here with a very small number of non-null queries.

\begin{table}[!ht]
\centering
\begin{tabular}{|c|c|c|c}
	\hline
	\multicolumn{3}{c}{\bf Summarizer}\\
	\hline
	\sc {\sc Cortex} & \textsc{\textsc{Artex}} & \sc Baseline  \\
	\hline
	 95   &\bf 950      	& 38   \\
	\hline
\end{tabular}
\caption{Number of queries having a non-null MAP.}
\label{tab:map2}
\end{table}

This low performance can be caused by the queries and references characteristics. 
We should investigate both properties effects on the summarization performance. 
Indeed according to our experiments Cortex was able to produce summaries informative with regard to the complete set of tweets which means it would be able to obtain greater performances over vague queries which have more than 20 relevant tweets.
Alternatively, Artex obtained interesting performance in selecting the most relevant tweets for queries having a limited number of tweets.

Another aspect is that the reference is limited since a tweet belongs to one and one only query. 
It then seems obvious since there are very similar queries. 

We also do not considered a decision strategy over all queries. 
That is to say, when a tweet appears to be very relevant to one query the system should avoid it when computing the scores for the others queries.
According to the way the reference is built, it would lead to better performances nevertheless, it will imply to drift from the original 'Search Problem'.

\section{Conclusive discussion}
\label{sec:conclu}

In this paper, we conducted extensive experiments on a real world dataset proposed by the CLEF-RepLab 2013 challenge.
One of the challenge proposed was to identify a group of tweets that can explain a particular query.
Such Information Retrieval or Search task can be evaluated as a traditional ranking information retrieval problem. 
In other words, systems will be a ranking of tweets where the most relevant appear at the first position of the returned list of tweets.

For each entity from RepLab we used the Topics proposed by E-watcher specialists as queries to build automatic summaries of the tweets steam.
These summaries reflect how does people perceive the entity with regard to this topic but also what do they mainly say about this topic.
We have presented rather simple statistical summarizers without knowledge which obtained interesting performances in this complex guided summarization task according to our evaluation.
The performed experiments demonstrate that the proposed methods are able to catch differences between tweets with regard to the queries.

Maybe more then summarizing social medias contents, evaluating the quality of these automatically generated summaries is an important research issue.
Besides keeping up-to-date with the last information, systems have to deal with contents informativity and to provide the most interesting tweets in order to help the end-user to answer his query.

Nevertheless, manually generating reference summaries for each possible query is not possible.
Moreover, as the issue is still relatively young, we are also lacking a specific automatic evaluation framework for this task.

The main contribution of this paper is finally to draw the base-line of query-based automatic summarization of Micro-Blogs and its evaluation. 
Our intuition is that by guessing the expected queries we are able to summarize the tweets stream and in this way ease a later clustering or classification stage.
Instead of working at a single tweet granularity, systems would be able to handle a summarized cluster.

In the future, we intend to extend the process by automatically generating summaries from the most informative tweets.
This way, for each query, we will allow the user to handle a short readable piece of text.

\section*{Acknowledgment}
This work was partly funded by the French National Research Agency (ANR), project ImagiWeb ANR-2012-CORD-002-01.

\bibliographystyle{plain}
\bibliography{bibtex}
\end{document}